\documentclass[runningheads]{llncs}
\usepackage{graphicx}
\usepackage{tikz}
\usepackage{comment}
\usepackage{amsmath,amssymb} % define this before the line numbering.
\usepackage{color}
\usepackage{subcaption}
\usepackage[font={small}]{caption}

\def\etal{\emph{et al}.}
\newcommand{\jc}[1]{\textcolor{black}{#1}}
\newcommand{\jcfinal}[1]{\textcolor{black}{#1}}
\newcommand{\mcfinal}[1]{\textcolor{black}{#1}}
\newcommand{\dg}[1]{\textcolor{black}{#1}}
\newcommand{\mc}[1]{\textcolor{black}{#1}}
\newcommand\blfootnote[1]{%
  \begingroup
  \renewcommand\thefootnote{}\footnote{#1}%
  \addtocounter{footnote}{-1}%
  \endgroup
}

\begin{document}
% \linenumbers
\pagestyle{headings}
\mainmatter
\def\ECCVSubNumber{3984}  % Insert your submission number here

\title{Learning Visual Context by Comparison} % Replace with your title

\begin{comment}
\titlerunning{ECCV-20 submission ID \ECCVSubNumber} 
\authorrunning{ECCV-20 submission ID \ECCVSubNumber} 
\author{Anonymous ECCV submission}
\institute{Paper ID \ECCVSubNumber}
\end{comment}
%******************

% \begin{comment}
\titlerunning{Learning Visual Context by Comparison}
\author{Minchul Kim*\inst{1} \and
Jongchan Park*\inst{1} \and
Seil Na\inst{1} \and\\
Chang Min Park\inst{2} \and
Donggeun Yoo$\dagger$\inst{1}}

\index{Park,Chang Min}
\authorrunning{M. Kim et al.}
\institute{Lunit Inc., Republic of Korea\\
\email{\{minchul.kim, jcpark, seil.na, dgyoo\}@lunit.io} \and
Seoul National University Hospital, Republic of Korea
\email{cmpark.morphius@gmail.com}}
% \end{comment}
%******************
\maketitle

\begin{abstract}
Finding diseases from an X-ray image is an important yet highly challenging task. 
Current methods for solving this task exploit various characteristics of the chest X-ray image, but one of the most important characteristics is still missing: the necessity of comparison between related regions in an image. 
In this paper, we present Attend-and-Compare Module (ACM) for capturing the difference between an object of interest and its corresponding context. 
We show that explicit difference modeling can be very helpful in tasks that require direct comparison between locations from afar.   
This module can be plugged into existing deep learning models.
For evaluation, we apply our module to three chest X-ray recognition tasks and COCO object detection \& segmentation tasks and observe consistent improvements across tasks. The code is available at \url{https://github.com/mk-minchul/attend-and-compare}.
% [JC]: NO \cite in the abstract, according to the guideline
\end{abstract}
\vspace{-4em}

\blfootnote{*The authors have equally contributed. $\dagger$Donggeun Yoo is the corresponding author.}
% introduction ===========================================
\section{Introduction}

%With the \mc{progress} in deep learning~\cite{lecun2015deep}, the medical image domain is receiving more attention~\cite{hinton2018deep} as there is much room for \mc{strengthening doctors' ability} \dg{with machine vision~\cite{nam2018development} \mc{in} their clinical practice. In fact, the recent developments of medical image recognition models have shown the potential for the continuous growth of diagnostic accuracy~\cite{paiva2017potential}.}
Among a variety of medical imaging modalities, chest X-ray is one of the most common and readily available examinations for diagnosing chest diseases. In the US, more than 35 million chest X-rays are taken every year~\cite{kamel2017utilization}. 
It is primarily used to screen diseases such as lung cancer, pneumonia, tuberculosis and pneumothorax to detect them at their earliest and most treatable stage.
However, the problem lies in the heavy workload of reading chest X-rays. Radiologists usually read tens or hundreds of X-rays every day.
% However, radiologists' workload of reading X-rays is increasing and many radiologists today have to read more than 100 X-ray studies every day. 
Several studies regarding radiologic errors~\cite{quekel1999miss,forrest1981radiologic} have reported that 20-30\% of exams are misdiagnosed.
% Hence, the accuracy of chest X-ray screening is low where 20-30\% of them are reported to be missed~\cite{quekel1999miss, forrest1981radiologic} and the need for more accurate computer-aided detection (CAD) tool to support radiologists is growing. 
To compensate for this shortcoming, many hospitals equip radiologists with computer-aided diagnosis systems.  
% \jc{To alleviate this problem, hospitals equip computer-aided diagnosis systems.}  
% The recent developments of medical image recognition models have shown potentials for growth in diagnostic accuracy~\cite{radiology2018}, and as a result, improving the performance of computer vision models in the medical image domain is an important topic. 
\jcfinal{The recent developments of medical image recognition models have shown potentials for growth in diagnostic accuracy~\cite{radiology2018}.}

With the recent presence of large-scale chest X-ray datasets~\cite{wang2017chestx,irvin2019chexpert,johnson2019mimic,bustos2019padchest}, there has been a long line of works that find thoracic diseases from chest X-rays using deep learning~\cite{yao2018weakly,guendel2018learning,Li_2018_CVPR,rajpurkar2017chexnet}.
Most of the works attempt to classify thoracic diseases, and some of the works further localize the lesions.
To improve recognition performance, Yao \etal~\cite{yao2018weakly} handles varying lesion sizes and Mao \etal~\cite{mao2019imagegcn} takes the relation between X-rays of the same patient into consideration. Wang \etal~\cite{wang2018chestnet} introduces an attention mechanism to focus on regions of diseases.
%~\cite{guan2018multi}는 설명이 없어서 뺐으니, 추후 넣어주세요~
% These works attempt to classify which diseases are in chest X-rays, and some try to pin-point abnormal locations using pixel-wise segmentation. 
% These works focused on exploiting several important characteristics in chest X-ray images such as diseases' variance in size~\cite{yao2018weakly}, relationships in multiple images taken by the same patient~\cite{mao2019imagegcn} or \jc{pathologies?}~\cite{guan2018multi}, and localization and attention mechanisms~\cite{wang2018chestnet}.

%KEEP) compare side to side, if on both sides the “nodules” in question are in the same position, then they are likely to be due to nipple shadows

\begin{figure}[t]
\centering
  \includegraphics[width=0.45\linewidth]{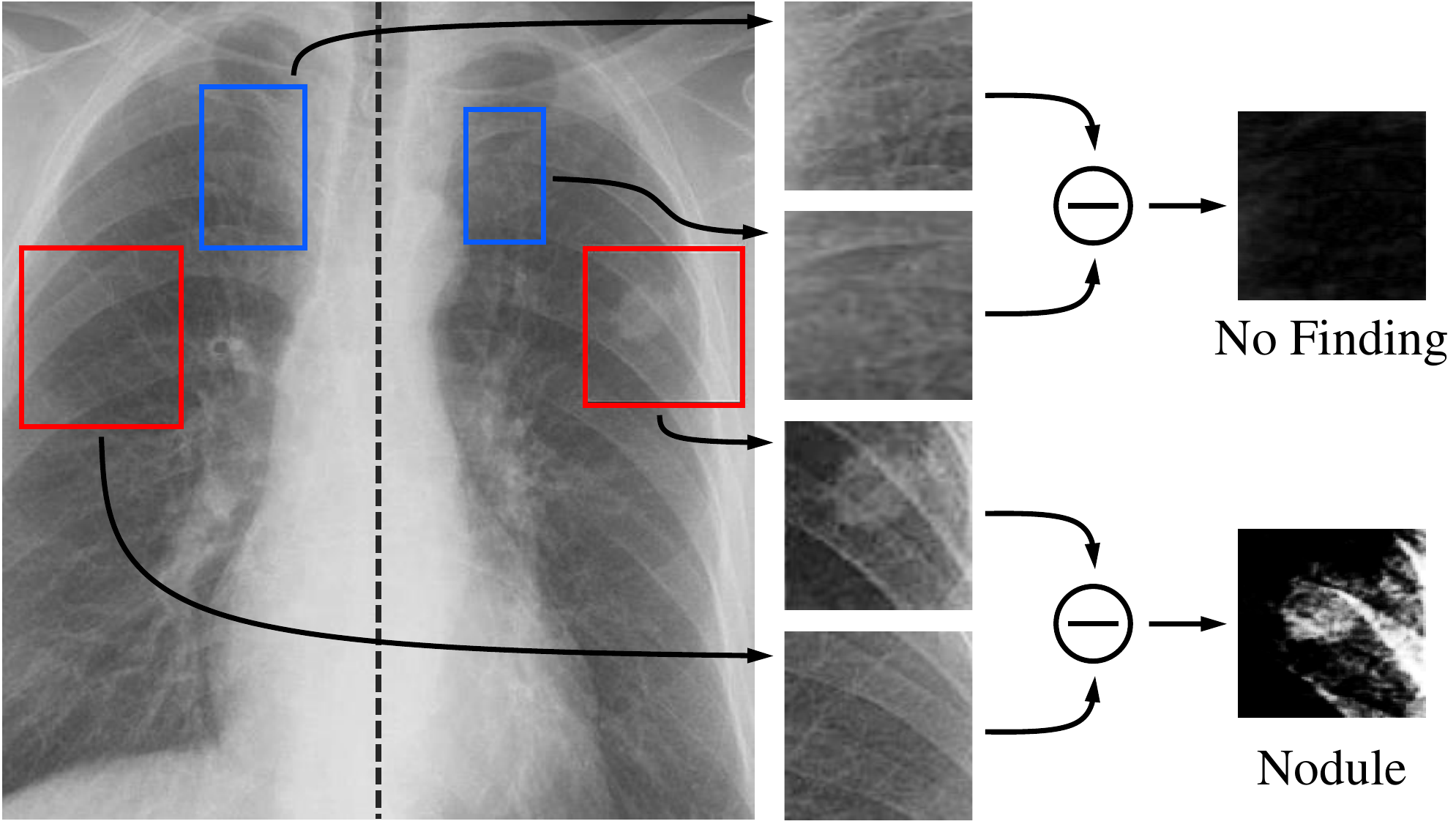}
  \caption{An example of a comparison procedure for radiologists. Little differences indicate no disease (blue), the significant difference is likely to be a lesion (red).}
  \label{figure:concept_figure}
\end{figure}

While these approaches were motivated by the characteristics of chest X-rays, we paid attention to how radiology residents are trained, which led to the following question: why don't we model the way radiologists read X-rays? When radiologists read chest X-rays, they \jcfinal{compare} zones~\cite{armato1994computerized}, paying close attention to any asymmetry between left and right lungs, or any changes between semantically related regions, that are likely to be due to diseases. 
\mcfinal{This comparison process provides contextual clues for the presence of a disease that local texture information may fail to highlight.}
% This comparison process provides important evidence of the presence of a disease and contributes to an accurate diagnosis. 
% \jc{This comparison process provides important context information for a more accurate diagnosis.}
\jc{Fig.~\ref{figure:concept_figure} illustrates an example of} the process. 
% \jcfinal{Considering the importance of context, previous studies~\cite{wang2018non,hu2018squeeze,cao2019gcnet,woo2018cbam} have proposed various context models, but none have addressed the need for the explicit procedure to \textit{compare} regions in an image.}
\mcfinal{Previous studies~\cite{wang2018non,hu2018squeeze,cao2019gcnet,woo2018cbam} proposed various context models, but none addressed the need for the explicit procedure to \textit{compare} regions in an image.}
% In visual recognition tasks with natural images, it has also been one of the major challenges to take visual contexts~\cite{wang2018non,hu2018squeeze,cao2019gcnet,woo2018cbam} into consideration. 
% The previous methods on contextual modeling aggregate the visual contexts from the feature to reflect them to the image representation. 
% However, none have previously addressed the need for utilizing the explicit procedure to compare different regions in an image.

In this paper, we present a novel module, called \textit{Attend-and-Compare Module} (ACM), that extracts features of an object of interest and a corresponding context to explicitly compare them by subtraction, mimicking the way radiologists read X-rays.
\jcfinal{Although motivated by radiologists' practices, we pose no explicit constraints for symmetry, and ACM learns to compare regions in a data-driven way.}
% ACM is a self-contained module that can be easily inserted into CNNs such as \mc{ResNet~\cite{he2016deep}, ResNeXt~\cite{xie2017aggregated} or DenseNet~\cite{huang2017densely}}. 
% Although initially motivated by the radiologic practice, ACM, according to our experiments, works well on natural image recognition tasks such as object detection and segmentation, implying that the difference information is an important aspect of context modeling.
%%% 앞서 비교를 통해 문제를 제기했는데, 여기서 비교가 다시 나오는게 흐름상 어색한것같아 주석처리합니다 -dg-
% 어디까지가 주석인가요? -mckim-
% \mc{HERE FIX NEEDED}
% Our module is inspired by several context modeling works such as non-local network~\cite{wang2018non}, squeeze-and-excitation network~\cite{hu2018squeeze}, GCNet~\cite{cao2019gcnet} and convolutional block attention module~\cite{woo2018cbam}. All of them are designed to model context in general computer vision tasks like image recognition,~\cite{imagenet_cvpr09}, object detection~\cite{lin2014microsoft} and action recognition~\cite{kay2017kinetics}, and we show that certain tasks such as detection in chest X-ray naturally requires comparison operation. Our context-aware module shows that explicitly modeling difference in features from afar can be beneficial in such scenarios.
% We validate ACM through visual recognition tasks including disease classification \& localization over three chest X-ray datasets~\cite{wang2017chestx} and object detection \& segmentation in COCO dataset~\cite{lin2014microsoft}. 
\jcfinal{ACM is validated over three chest X-ray datasets~\cite{wang2017chestx} and object detection \& segmentation in COCO dataset~\cite{lin2014microsoft} with various backbones such as ResNet~\cite{he2016deep}, ResNeXt~\cite{xie2017aggregated} or DenseNet~\cite{huang2017densely}.} 
% The experimental results on chest X-ray datasets demonstrate that the explicit comparison process by ACM indeed improves the disease classification and lesion localization performance. 
% \jcfinal{The experimental results on chest X-ray datasets demonstrate that the explicit comparison process by ACM indeed improves the recongition performance.} 
% On COCO dataset, the object detection and segmentation model equipped with ACM also significantly outperforms the previous modules designed for context modeling.
% \jc{
% Although initially motivated by the radiologic practice, ACM works surprisingly well on natural images. In COCO dataset, ACM outperforms or is competitive with previous works on context modeling, implying that attend-and-compare is a useful practice in natural image domain}.
%\jc{Although initially motivated by the radiologic practice, ACM works surprisingly well in COCO dataset, implying that comparison process is useful in natural images}.
% Although ACM stems from radiologic practice, it also performs well on natural image domain such as COCO dataset.
\mcfinal{Experimental results on chest X-ray datasets and natural image datasets demonstrate that the explicit comparison process by ACM indeed improves the recognition performance.}

\subsubsection{Contributions}
To sum up, our major contributions are \dg{as follows}\jc{:}
\begin{enumerate}
    \item \dg{We propose} a novel context module called ACM that explicitly compares different regions, following the way radiologists read chest X-rays.
    \item \dg{The proposed ACM captures multiple comparative self-attentions whose difference is beneficial to recognition tasks. }
    % \item \jc{We demonstrate the effectiveness of ACM on three chest X-ray datasets~\cite{wang2017chestx} and COCO detection \& segmentation dataset~\cite{lin2014microsoft}. ACM shows consistent improvements in various architectures and superior performance over previous works across all datasets.}
    \item We demonstrate the effectiveness of ACM on three chest X-ray datasets~\cite{wang2017chestx} and COCO detection \& segmentation dataset~\cite{lin2014microsoft} \jcfinal{with various architectures}.%\jcfinal{ACM shows superior performance over previous works across all tasks with various architectures.}
\end{enumerate}

%------------------------------------------------------------------------
\section{Related Work}

\subsection{Context Modeling}
Context modeling in deep learning is primarily conducted with the self-attention mechanism~\cite{vaswani2017attention,hu2018squeeze,lee2019srm,roy2018concurrent}. Attention related works are broad and some works do not explicitly pose themselves in the frame of context modeling. 
However, we include them to highlight different methods that make use of global information, which can be viewed as context.

In the visual recognition domain, recent self-attention mechanisms~\cite{hu2018squeeze,wang2017residual,chen2017sca,lee2019srm,woo2018cbam} generate dynamic attention maps for recalibration (e.g., emphasize salient regions or channels).
Squeeze-and-Excitation network (SE)~\cite{hu2018squeeze} learns to model channel-wise attention using the spatially averaged feature. 
A Style-based Recalibration Module (SRM)~\cite{lee2019srm} further explores the global feature modeling in terms of style recalibration. 
Convolutional block attention module (CBAM)~\cite{woo2018cbam} extends SE module to the spatial dimension by sequentially attending the important location and channel given the feature. 
The attention values are computed with global or larger receptive fields, and thus, more contextual information can be embedded in the features. 
However, as the information is aggregated into a single feature by average or similar operations, spatial information from the relationship among multiple locations may be lost.
% However, the drawback of such mechanism comes from the loss in information during the creation of averaged global feature, which cannot fully capture the relationship among multiple locations.

Works that explicitly tackle the problem of using context stem from using pixel-level pairwise relationships~\cite{wang2018non,huang2019ccnet,cao2019gcnet}. 
Such works focus on long-range dependencies and explicitly model the context aggregation from dynamically chosen locations.
Non-local neural networks (NL)~\cite{wang2018non} calculate pixel-level pairwise relationship weights and aggregate (weighted average) the features from all locations according to the weights. 
The pairwise modeling may represent a more diverse relationship, but it is computationally more expensive. 
As a result, Global-Context network (GC)~\cite{cao2019gcnet} challenges the necessity of using all pairwise relationships in NL and suggests to softly aggregate a single distinctive context feature for all locations. 
Criss-cross attention (CC)~\cite{huang2019ccnet} for semantic segmentation reduces the computation cost of NL by replacing the pairwise relationship attention maps with criss-cross attention block which considers only horizontal and vertical directions separately.
NL and CC explicitly model the pairwise relationship between regions with affinity metrics, but the qualitative results in~\cite{wang2018non,huang2019ccnet} demonstrate a tendency to aggregate features only among foreground objects or among pixels with similar semantics. 

\jc{Sharing a similar philosophy, there have been works on contrastive attention~\cite{song2018mask,zhang2018top}. MGCAM~\cite{song2018mask} uses the contrastive feature between persons and backgrounds, but it requires extra mask supervision for persons. C-MWP~\cite{zhang2018top} is a technique for generating more accurate localization maps in a contrastive manner, but it is not a learning-based method and uses pretrained models.}

\begin{figure*}[t]
    \centering
    \includegraphics[width=\linewidth]{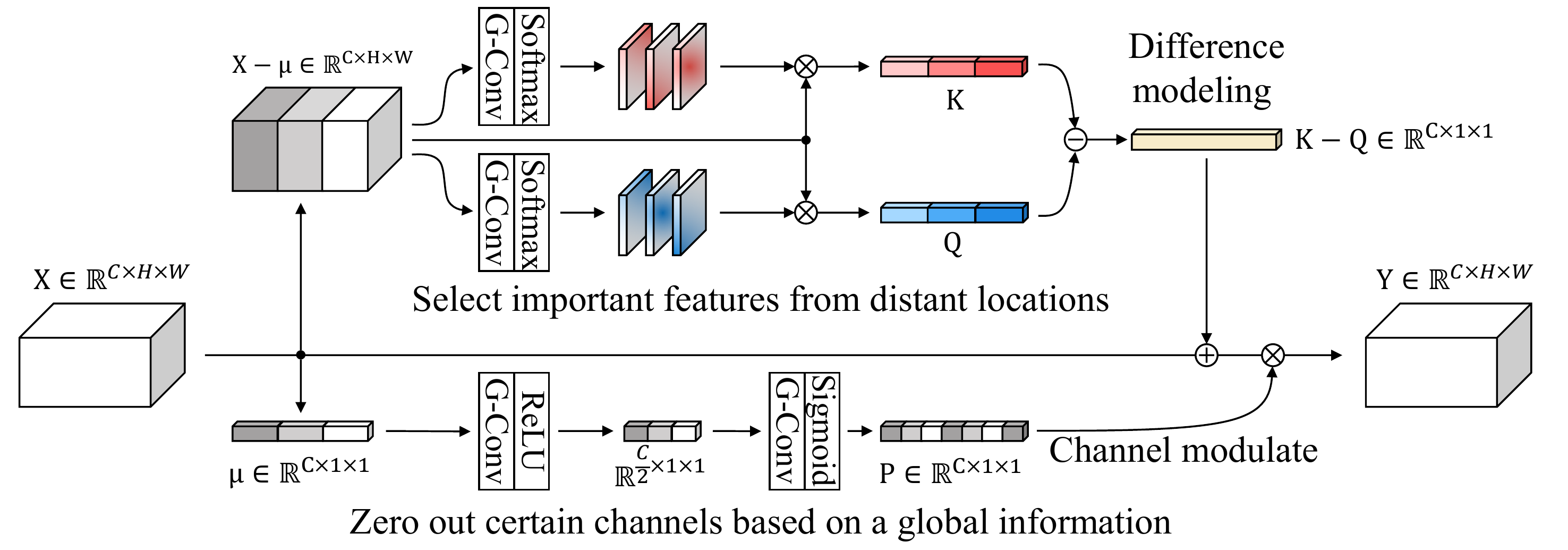}
    \caption{\dg{Illustration of the ACM module.} It takes in an input feature and uses the mean\dg{-}subtracted feature to calculate two feature vectors ($K$, $Q$). \mc{Each feature vector ($K$ or $Q$) contains multiple attention vectors from multiple locations calculated using grouped convolutions and normalizations.} The difference \dg{of the vectors} is added to the main feature to make the information more distinguishable. The resulting feature is modulated channel-wise, by the global information feature.}
    \label{figure:architecture}
\end{figure*}

Inspired by how radiologists diagnose, our proposed module, namely, ACM explicitly models a comparing mechanism. The overview of the module can be found in Fig.~\ref{figure:architecture}.
Unlike the previous works proposed in the natural image domain, our work stems from the precise need for incorporating difference operation in reading chest radiographs. \dg{Instead of} \mc{finding an affinity map based attention as in NL~\cite{wang2018non}, ACM explicitly uses direct comparison procedure for context modeling; instead of using extra supervision to localize regions to compare as in MGCAM~\cite{song2018mask}, ACM automatically learns to focus on meaningful regions to compare. Importantly, the efficacy of our explicit and data-driven contrastive modeling is shown by the superior performance over other context modeling works.}

\subsection{Chest X-ray as a Context-Dependent Task}

Recent releases of the large-scale chest X-ray datasets~\cite{wang2017chestx,irvin2019chexpert,johnson2019mimic,bustos2019padchest} showed that commonly occurring diseases can be classified and located in a weakly-supervised multi-label classification framework. 
ResNet~\cite{he2016deep} and DenseNet~\cite{huang2017densely,rajpurkar2017chexnet} pretrained on ImageNet~\cite{imagenet_cvpr09} have set a strong baseline for these tasks, and other studies have been conducted on top of them to cover various issues of recognition task in the chest X-ray modality.

To address the issue of localizing diseases using only class-level labels, Guendel \etal~\cite{guendel2018learning} propose an auxiliary localization task where the ground truth of the location of the diseases is extracted from the text report. 
Other works use attention module to indirectly align the class-level prediction with the potentially abnormal location~\cite{wang2018chestnet,tang2018attention,guan2018multi} without the text reports on the location of the disease.
Some works observe that although getting annotations for chest X-rays is costly, it is still helpful to leverage both a small number of location annotations and a large number of class-level labels to improve both localization and classification performances~\cite{Li_2018_CVPR}. Guan~\etal~\cite{guan2018diagnose} also proposes a hierarchical hard-attention for cascaded inference. 

In addition to such characteristics inherent in the chest X-ray image, we would like to point out that the difference between an object of interest and a corresponding context could be the crucial key for classifying or localizing several diseases as it is important to compare semantically meaningful locations. 
However, despite the importance of capturing the semantic difference between regions in chest X-ray recognition tasks, no work has dealt with it yet. 
Our work is, to the best of our knowledge, the first to utilize this characteristic in the Chest X-ray image recognition setting.

\section{Attend-and-Compare Module}\label{sec:module}
\subsection{Overview}
Attend-and-Compare Module (ACM) extracts an object of interest and the corresponding context to compare, and enhances the original image feature with the comparison result. Also, ACM is designed to be light-weight, self-contained, and \mc{compatible with popular backbone architectures~\cite{he2016deep,huang2017densely,xie2017aggregated}.} We formulate ACM comprising three procedures as
\begin{equation}
Y = f_\text{ACM}(X) = P(X+(K-Q)),
\label{eq:acm}
\end{equation}
where $f_\text{ACM}$ is a transformation mapping an input feature $X\in \mathbb{R}^{C\times H \times W}$ to an output feature $Y\in \mathbb{R}^{C\times H \times W}$ in the same dimension. Between $K\in\mathbb{R}^{C\times1\times1}$ and $Q\in \mathbb{R}^{C\times1\times1}$, one is intended to be the object of interest and the other is the corresponding context. ACM compares the two by subtracting one from the other, and add the comparison result to the original feature $X$, followed by an additional channel re-calibration operation with $P\in \mathbb{R}^{C\times1\times1}$. Fig.~\ref{figure:architecture} illustrates Equation~(\ref{eq:acm}). These three features $K$, $Q$ and $P$ are conditioned on the input feature $X$ and will be explained in details below.

\subsection{Components of ACM}
\subsubsection{Object of Interest and Corresponding Context}
To fully express the relationship between different spatial regions of an image, ACM generates two features ($K, Q$) that focus on two spatial regions of the input feature map $X$. At first, ACM normalizes the input feature map as $X:=X-\mu$ where $\mu$ is a $C$-dimensional mean vector of $X$. We include this procedure to make training more stable as $K$ and $Q$ will be generated by learnable parameters ($W_K, W_Q$) that are shared by all input features. Once $X$ is normalized, ACM then calculates $K$ with $W_K$ as
\begin{equation}
K = \sum_{i,j \in H,W} \frac{\mbox{exp}(W_K X_{i,j})}{\sum_{H,W} \mbox{exp}(W_K X_{h,w})} X_{i,j},
\end{equation}
where $X_{i,j}\in\mathbb{R}^{C\times1\times1}$ is a vector at a spatial location ($i,j$) and $W_K\in\mathbb{R}^{C\times1\times1}$ is a weight of a $1\times1$ convolution. The above operation could be viewed as applying $1\times 1$ convolution on the feature map $X$ to obtain a single-channel attention map in $\mathbb{R}^{1\times H\times W}$, applying softmax to normalize the attention map, and finally weighted averaging the feature map $X$ using the normalized map. $Q$ is also modeled likewise, but with $W_Q$. $K$ and $Q$ serve as features representing important regions in $X$. % \dg{[subtraction and addition of X+(K-Q) should be shortly explained here before introducing channel recalibration]}
\mc{We add $K-Q$ to the original feature so that the comparative information is more distinguishable in the feature.}

\subsubsection{Channel Re-calibration}
In light of the recent success in self-attention modules which use a globally pooled feature to re-calibrate channels~\cite{hu2018squeeze,lee2019srm,woo2018cbam}, we calculate the channel re-calibrating feature $P$ as
\begin{equation}
P=\sigma\circ\text{conv}_2^{1\times 1}\circ\text{ReLU}\circ\text{conv}_1^{1\times 1}(\mu),
%P = \sigma\left(\text{conv}_{1\times 1} \left(\delta\left(\text{conv}_{1\times 1}\left(\mu\right)\right)\right)\right),
\end{equation}
where $\sigma$ and $\text{conv}^{1\times1}$ denote a sigmoid function and a learnable $1\times1$ convolution function, respectively. The resulting feature vector $P$ will be multiplied to $X+(K-Q)$ to scale down certain channels. $P$ can be viewed as marking which channels to attend with respect to the task at hand.

\subsubsection{Group Operation}
To model a relation of multiple regions from a single module, we choose to incorporate group-wise operation. 
We replace all convolution operations with grouped convolutions~\cite{krizhevsky2012imagenet,xie2017aggregated}, where the input and the output are divided into $G$ number of groups channel-wise, and convolution operations are performed for each group separately. 
In our work, we use the grouped convolution to deliberately represent multiple important locations from the input.
\jc{Here, we compute $G$ different attention maps by applying grouped convolution to $X$, and then obtain the representation $K = [K^1,\cdots,K^G]$ by aggregating each group in X with each attention as follows:}
\begin{equation}
K^g=\sum_{i,j\in H,W}\frac{\mbox{exp}(W^g_K X^g_{i,j})}{\sum_{H,W}\mbox{exp}(W^g_K X^g_{h,w})} X^g_{i,j},
\end{equation}
\noindent where $g$ refers to $g$-th group. 

\begin{comment}
\begin{equation}
    K = \sum_{H\times W}\left(\underset{C\times H\times W}{X} \circ \underset{G\times H\times W}{A}\right),\,\,\,\,\, A = \underset{ H,W}{\text{SoftMax}}\left[ \text{GroupConv}_{C\rightarrow G}(X)\right]
\end{equation}
\end{comment}

\subsubsection{Loss Function}
ACM learns to utilize comparing information within an image by modeling $\{K, Q\}$ whose difference can be important for the given task. To further ensure diversity between them, we introduce an orthogonal loss. based on a dot product.  It is defined as
\begin{equation}
\ell_\text{orth}(K, Q)=\dfrac{K\cdot Q}{C},
\end{equation}
where $C$ refers to the number of channels. Minimizing this loss can be viewed as decreasing the similarity between $K$ and $Q$. One trivial solution to minimizing the term would be making $K$ or $Q$ zeros, but they cannot be zeros as they come from the weighted averages of $X$. The final loss function for a target task can be written as
\begin{equation}
\ell_\text{task}+\lambda\sum_{m}^M\ell_{\text{orth}}(K_m, Q_m),
\end{equation}
where $\ell_\text{task}$ refers to a loss for the target task, and $M$ refers to the number of ACMs inserted into the network. $\lambda$ is a constant for controlling the effect of the orthogonal constraint.

\subsubsection{Placement of ACMs}
In order to model contextual information in various levels of feature representation, we insert multiple ACMs into the backbone network. In ResNet, following the placement rule of SE module~\cite{hu2018squeeze}, we insert the module at the end of every Bottleneck block. For example, a total of 16 ACMs are inserted in ResNet-50. \mc{Since DenseNet contains more number of DenseBlocks than ResNet's Bottleneck block, we inserted ACM in DenseNet every other three DenseBlocks.} \jcfinal{Note that we did not optimize the placement location or the number of placement for each task.} While we use multiple ACMs, the use of grouped convolution significantly reduces the computation cost in each module.

\section{Experiments}\label{sec:exp}

We evaluate ACM in several datasets: internally-sourced Emergency-Pneumothorax (Em-Ptx) and Nodule (Ndl) datasets for lesion localization in chest X-rays, Chest X-ray14~\cite{wang2017chestx} dataset for multi-label classification, and COCO 2017~\cite{lin2014microsoft} dataset for object detection and instance segmentation. 
% Note that the internally-sourced datasets will not be disclosed due to patients' privacy issues. 
The experimental results show that ACM outperforms other context-related modules, in both chest X-ray tasks and natural image tasks. 

\subsubsection{Experimental Setting} Following the previous study~\cite{baltruschat2019comparison} on multi-label classification with chest X-Rays, we mainly adopt ResNet-50 as our backbone network. To show generality, we sometimes adopt DenseNet~\cite{huang2017densely} and ResNeXt~\cite{xie2017aggregated} as backbone networks.
In classification tasks, we report class-wise Area Under the Receiver Operating Characteristics (AUC-ROC) for classification performances. 
For localization tasks, we report the jackknife free-response receiver operating characteristic (JAFROC)~\cite{chakraborty2005recent} for localization performances. JAFROC is a metric widely used for tracking localization performance in radiology. 
All chest X-ray tasks are a weakly-supervised setting~\cite{oquab2015object} in which the model outputs a probability map for each disease, and final classification scores are computed by global maximum or average pooling.
If any segmentation annotation is available, extra map losses are given on the class-wise confidence maps. 
For all experiments, we initialize the backbone weights with ImageNet-pretrained weights and randomly initialize context-related modules. 
% \jcfinal{In order to avoid trivial confounding relationships, the internal chest X-ray datasets (Em-Ptx, Ndl) are sub-sampled from a pool of real-world cohort. In the cohort, 10 major x-ray findings (Nodule, Consolidation, etc) and various medical devices (EKG, Endotracheal tube, Chemoport, etc) are present.}
Experiment details on each dataset are elaborated in the following section.

\subsection{Localization on Em-Ptx Dataset}\label{exp:em_ptx}
% task introduction
\subsubsection{Task Overview} The goal of this task is to localize emergency-pneumothorax (Em-Ptx) regions. 
Pneumothorax is a fatal thoracic disease that needs to be treated immediately. 
As a treatment, a medical tube is inserted into the pneumothorax affected lung. 
It is often the case that a treated patient repeatedly takes chest X-rays over a short period to see the progress of the treatment. 
Therefore, a chest X-ray with pneumothorax is categorized as an emergency,  but a chest X-ray with both pneumothorax and a tube is not an emergency. 
The goal of the task is to correctly classify and localize emergency-pneumothorax. 
To accurately classify the emergency-pneumothorax, the model should exploit the relationship between pneumothorax and tube within an image, even when they are far apart. 
In this task, utilizing the context as the presence/absence of a tube is the key to accurate \mc{classification.}

\begin{figure}[t]
\small
\centering
\begin{tabular}{cc}
\includegraphics[height=2.8cm]{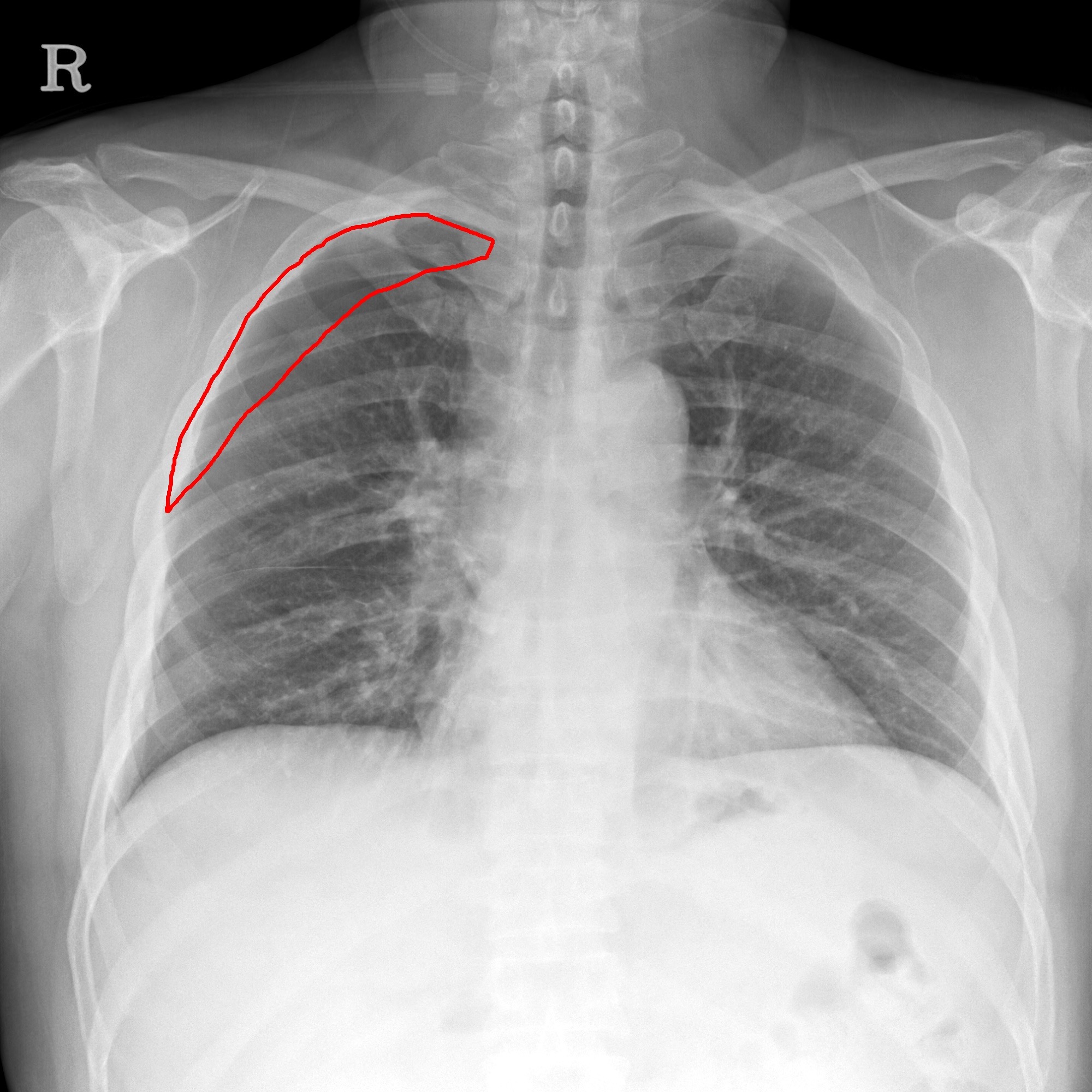}&
\includegraphics[height=2.8cm]{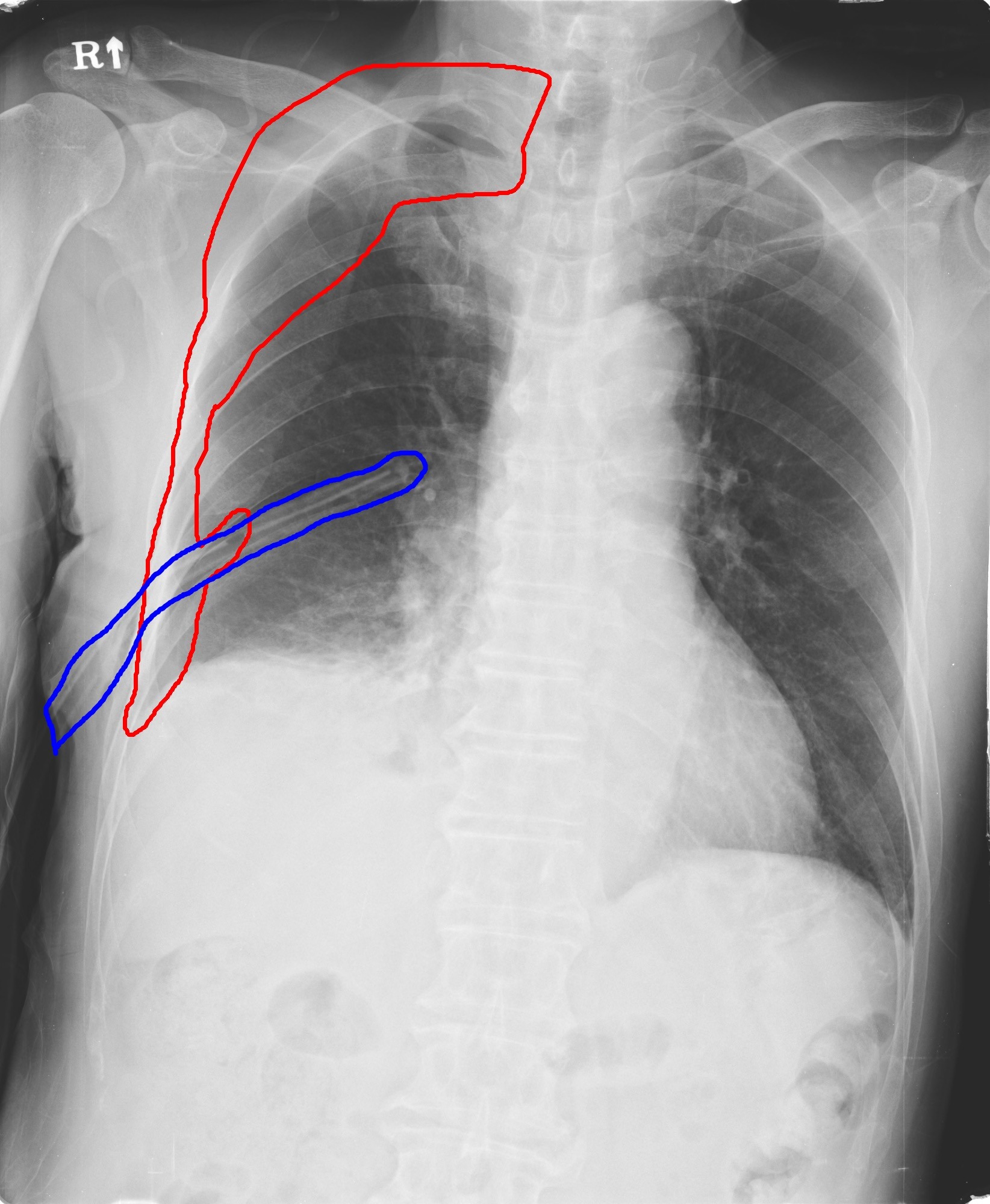}\\
(a) Emergency &(b) Non-emergency
\end{tabular}
\caption{\jc{Examples of pneumothorax cases and annotation maps in Em-Ptx dataset. Lesions are drawn in red. (a) shows a case with pneumothorax, and (b) shows a case which is already treated with a medical tube marked as blue.}}
\label{figure:ptx_example}
\end{figure}

% dataset composition
We internally collected 8,223 chest X-rays, including 5,606 pneumothorax cases, of which 3,084 cases are emergency. 
\mcfinal{The dataset is from a real-world cohort and contains cases with other abnormalities even if they do not have pneumothorax. We received annotations for 10 major x-ray findings (Nodule, Consolidation, etc) and the presence of medical devices (EKG, Endotracheal tube, Chemoport, etc). Their labels were not used for training.}
The task is a binary classification and localization of emergency-pneumothorax. 
All cases with pneumothorax and tube together and all cases without pneumothorax are considered a non-emergency. 
\jc{We separated 3,223 cases as test data, of which 1,574 cases are emergency-pneumothorax, 1,007 cases are non-emergency-pneumothorax, and 642 cases are pure normal. 
All of the 1,574 emergency cases in the test data were annotated with coarse segmentation maps by board-certified radiologists. }
Of the 1,510 emergency-pneumothorax cases in the training data, only 930 cases were annotated, and the rest were used with only the class label.
An example of a pneumothorax case and an annotation map created by board-certified radiologists is provided in Fig. \ref{figure:ptx_example}.

\subsubsection{Training Details}
We use Binary Cross-Entropy loss for both classification and localization, and SGD optimizer with momentum 0.9.  The initial learning rate is set to 0.01. The model is trained for 35 epochs in total and the learning rate is dropped by the factor of 10 at epoch 30. For each experiment setting, we report the average AUC-ROC and JAFROC of 5 runs with different initialization.

\subsubsection{Result}
The experimental result is summarized in Table~\ref{table:em_ptx}.
Compared to the baseline ResNet-50 and ResNet-101, all the context modules have shown performance improvements.
ACM outperforms all other modules in terms of both AUC-ROC and JAFROC.
The result supports our claim that the contextual information is critical to Em-Ptx task, and our module, with its explicit feature-comparing design, shows the biggest improvement in terms of classification and localization.

\begin{table}[t]
\centering
\caption{Results on Em-Ptx dataset. Average of 5 random runs are reported for each setting with standard deviation. RN stands for ResNet~\cite{he2016deep}.}
\begin{tabular}{|l|r|r|l|r|r|}
\hline
Method & AUC-ROC& JAFROC & Method & AUC-ROC & JAFROC\\
\hline\hline
RN-50 & 86.78\(\pm\)0.58 & 81.84\(\pm\)0.64
& RN-101&89.75\(\pm\)0.49&85.36\(\pm\)0.44\\
RN-50 + SE~\cite{hu2018squeeze} & 93.05\(\pm\)3.63 & 89.19\(\pm\)4.38
& RN-101 + SE~\cite{hu2018squeeze}&90.36\(\pm\)0.83&85.54\(\pm\)0.85\\
% R50 + SRM~\cite{lee2019srm} & 0.9461\(\pm\)0.0053 & 0.9189\(\pm\)0.0069 \\
RN-50 + NL~\cite{wang2018non} & 94.63\(\pm\)0.39 & 91.93\(\pm\)0.68
& RN-101 + NL~\cite{wang2018non}&94.24\(\pm\)0.34&91.70\(\pm\)0.83\\
RN-50 + CC~\cite{huang2019ccnet} & 87.73\(\pm\)8.66 & 83.32\(\pm\)10.36
& RN-101 + CC~\cite{huang2019ccnet}&92.57\(\pm\)0.89&89.75\(\pm\)0.89\\
RN-50 + ACM & \textbf{95.35}\(\pm\)0.12 & \textbf{94.16}\(\pm\)0.21
& RN-101 + ACM&\textbf{95.43}\(\pm\)0.14&\textbf{94.47}\(\pm\)0.10\\
\hline
\end{tabular}
\label{table:em_ptx}
\end{table}

\begin{table}[t]
\centering
\caption{Performance with respect to varying module architectures and hyper-parameters on Em-Ptx dataset. All the experiments are based on ResNet-50.}
    \begin{subtable}{0.55\textwidth}
    \centering
        \begin{tabular}{|l|r|r|}
        \hline
        Module&AUC-ROC &JAFROC \\
        \hline\hline
        None&86.78\(\pm\)0.58&81.84\(\pm\)0.64\\
        $X+(K-Q)$&94.25\(\pm\)0.31&92.94\(\pm\)0.36\\
        $PX$&87.16\(\pm\)0.42&82.05\(\pm\)0.30\\
        $P(X+K)$&94.96\(\pm\)0.15&93.59\(\pm\)0.24\\
        $P(X+(K-Q))$&\textbf{95.35}\(\pm\)0.12&\textbf{94.16}\(\pm\)0.21\\
        \hline
        \multicolumn{3}{c}{(a) Ablations on $K,Q$ and $P$.}
        \end{tabular}
    \end{subtable}%
    % \begin{subtable}{0.5\textwidth}
    % \centering
    %     \begin{tabular}{|l|r|r|}
    %     \hline
    %     Module&AUC-ROC &JAFROC \\
    %     \hline\hline
    %     None&86.78\(\pm\)0.58&81.84\(\pm\)0.64\\
    %     $PX$&87.16\(\pm\)0.42&82.05\(\pm\)0.30\\
    %     % $X+(K-Q)$&94.25\(\pm\)0.31&92.94\(\pm\)0.36\\
    %     $P(X+K)$&94.96\(\pm\)0.15&93.59\(\pm\)0.24\\
    %     $P(X+(K-Q))$&\textbf{95.35}\(\pm\)0.12&\textbf{94.16}\(\pm\)0.21\\
    %     \hline
    %     \multicolumn{3}{c}{(b) Ablations on $K,Q$ and $P$.}
    %     \end{tabular}
    % \end{subtable}
    \begin{subtable}{0.4\textwidth}
    \centering
        \begin{tabular}{|l|r|r|}
        \hline
        \#groups&AUC-ROC&JAFROC\\
        \hline\hline
        8&90.96\(\pm\)1.88&88.79\(\pm\)2.23\\
        32&\textbf{95.35}\(\pm\)0.12&\textbf{94.16}\(\pm\)0.21\\
        64&95.08\(\pm\)0.25&93.73\(\pm\)0.31\\
        128&94.89\(\pm\)0.53&92.88\(\pm\)0.53\\
        \hline
        \multicolumn{3}{c}{(b) Ablations on number of groups.}
        \end{tabular}
    \end{subtable}
    \begin{subtable}{0.6\textwidth}
    \centering
        \begin{tabular}{|l|r|r|}
        \hline
        $\lambda$&AUC-ROC&JAFROC\\
        \hline\hline
        0.00&95.11\(\pm\)0.20&93.87\(\pm\)0.20\\
        0.01&95.29\(\pm\)0.34&94.09\(\pm\)0.41\\
        0.10&\textbf{95.35}\(\pm\)0.12&\textbf{94.16}\(\pm\)0.21\\
        1.00&95.30\(\pm\)0.17&94.04\(\pm\)0.11\\
        \hline
        \multicolumn{3}{c}{(c) Ablations on orthogonal loss weight $\lambda$.}
        \end{tabular}
    \end{subtable}%
    
\label{table:analysis}
\end{table}

\subsection{Analysis on ACM with Em-Ptx dataset}\label{exp:aisblation}
In this section, we empirically validate the efficacy of each component in ACM and search for the optimal hyperparameters.
For the analysis, we use the Em-Ptx dataset. The training setting is identical to the one used in Sec.~\ref{exp:em_ptx}.
The average of 5-runs is reported.

\subsubsection{Effect of Sub-modules}
As described in Sec.~\ref{sec:module}, our module consists of 2 sub-modules: difference modeling and channel modulation.
We experiment with the two sub-modules both separately and together.
The results are shown in Table~\ref{table:analysis}.
Each sub-module brings improvements over the baseline, indicating the context modeling in each sub-module is beneficial to the given task. 
Combining the sub-modules brings extra improvement, showing the complementary effect of the two sub-modules.
The best performance is achieved when all sub-modules are used.

\subsubsection{Number of Groups}
By dividing the features into groups, the module can learn to focus on multiple regions, with only negligible computational or parametric costs. 
On the other hand, too many groups can result in too few channels per group, which prevents the module from exploiting correlation among many channels. We empirically find the best setting for the number of groups. The results are summarized in Table~\ref{table:analysis}.
Note that, training diverges when the number of groups is 1 or 4.
The performance improves with the increasing number of groups and saturates after 32.
We set the number of groups as 32 across all other experiments, \mc{except in DenseNet121 \dg{(Sec.~\ref{exp:cxr14})} where we set the number of channel per group to 16 due to channel divisibility.}

\subsubsection{Orthogonal Loss Weight}
We introduce a new hyperparameter \(\lambda\) to balance between the target task loss and the orthogonal loss.
Although the purpose of the orthogonal loss is to diversify the compared features, an excessive amount of \(\lambda\) can disrupt the trained representations.
\jcfinal{
We empirically determine the magnitude of \(\lambda\). Results are summarized in Table~\ref{table:analysis}.
% \(\lambda\) with value 0.0 indicates that the model is trained without the orthogonal loss.
From the results, we can empirically verify the advantageous effect of the orthogonal loss on the performance, and the optimal value of \(\lambda\) is 0.1.}
\jcfinal{In the validation set, the average absolute similarities between K and Q with \(\lambda=0,0.1\) are 0.1113 and 0.0394, respectively. It implies that K and Q are dissimilar to some extent, but the orthogonal loss further encourages it.}
We set \(\lambda\) as 0.1 across all other experiments.

\subsection{Localization on Ndl Dataset}\label{exp:ndl}
% task introduction
\subsubsection{Task Overview} The goal of this task is to localize lung nodules (Ndl) in chest X-ray images. 
Lung nodules are composed of fast-growing dense tissues and thus are displayed as tiny opaque regions. 
% The challenge is the inter-patient variability, view-point changes and differences in imaging devices. 
Due to inter-patient variability, view-point changes and differences in imaging devices, 
the model that learns to find nodular patterns with respect to the normal side of the lung from the same image (context) may generalize better.
We collected 23,869 X-ray images, of which 3,052 cases are separated for testing purposes. 
Of the 20,817 training cases, 5,817 cases have nodule(s). 
Of the 3,052 test cases, 694 cases are with nodules. 
Images without nodules may or may not contain other lung diseases. 
All cases with nodule(s) are annotated with coarse segmentation maps by board-certified radiologists. 
\jcfinal{We use the same training procedure for the nodule localization task as for the Em-Ptx localization. 
We train for 25 epochs with the learning rate dropping once at epoch 20 by the factor of 10.}

% \subsubsection{Training Details} 
% We use the same training procedure for the nodule localization task as for the Em-Ptx localization. 
% We train for 25 epochs with the learning rate dropping once at epoch 20 by the factor of 10.

\subsubsection{Results} 
The experimental result is summarized in Table~\ref{table:nodule}.
ACM outperforms all other context modeling methods in terms of both AUC-ROC and JAFROC. 
The results support our claim that the comparing operation, motivated by how radiologists read X-rays, provides a good contextual representation that helps with classifying and localizing lesions in X-rays. Note that the improvements in this dataset may seem smaller than in the Em-Ptx dataset. \mc{In the usage of context modules in general, the bigger increase in performance in Em-Ptx dataset is because emergency classification requires knowing both the presence of the tube and the presence of Ptx even if they are far apart. So the benefit of the contextual information is directly related to the performance. However, nodule classification can be done to a certain degree without contextual information. Since not all cases need contextual information, the performance gain may be smaller.}

\begin{table}[t]
\centering
\small
\setlength{\tabcolsep}{3pt}
%\begin{center}
\caption{Results on Ndl dataset. Average of 5 random runs are reported for each setting with standard deviation.}
\begin{tabular}{|l|r|r|}
\hline
Method & AUC-ROC & JAFROC \\
\hline\hline
ResNet-50 & 87.34\(\pm\)0.34 & 77.35\(\pm\)0.50 \\
ResNet-50 + SE~\cite{hu2018squeeze} & 87.66\(\pm\)0.40 & 77.57\(\pm\)0.44 \\
% ResNet-50 + SRM~\cite{lee2019srm} & 0.9461\(\pm\)0.0053 & 0.9189\(\pm\)0.0069 \\
ResNet-50 + NL~\cite{wang2018non} & 88.35\(\pm\)0.35 & 80.51\(\pm\)0.56 \\
ResNet-50 + CC~\cite{huang2019ccnet} & 87.72\(\pm\)0.18 & 78.63\(\pm\)0.40 \\
ResNet-50 + ACM & \textbf{88.60}\(\pm\)0.23 & \textbf{83.03}\(\pm\)0.24 \\
\hline
\end{tabular}

\label{table:nodule}
\end{table}

\subsection{Multi-label Classification on Chest X-ray14}\label{exp:cxr14}
\subsubsection{Task Overview} In this task, the objective is to identify the presence of 14 diseases in a given chest X-ray image.
Chest X-ray14~\cite{wang2017chestx} dataset is the first large-scale dataset on 14 common diseases in chest X-rays. 
It is used as a benchmark dataset in previous studies~\cite{yao2018weakly,guendel2018learning,rajpurkar2017chexnet,chen2019dualchexnet,mao2019imagegcn,Li_2018_CVPR,tang2018attention,guan2018multi}.
The dataset contains 112,120 images from 30,805 unique patients. 
Image-level labels are mined from image-attached reports using natural language processing techniques (each image can have multi-labels). We split the dataset into training (70\%), validation (10\%), and test (20\%) sets, following previous works~\cite{wang2017chestx,yao2018weakly}.

\subsubsection{Training Details} \mc{As shown in Table \ref{table:cxr14}, previous works on CXR14 dataset vary in loss, input size, etc. 
We use CheXNet~\cite{rajpurkar2017chexnet} implementation\footnote{\url{https://github.com/jrzech/reproduce-chexnet}} to conduct context module comparisons, and varied with the backbone architecture and the input size to find if context modules work in various settings.} 
\mc{We use BCE loss with the SGD optimizer with momentum 0.9 and weight decay 0.0001.} 
Although ChexNet uses the input size of 224, we use the input size of 448 as it shows a better result than 224 with DenseNet121.
\jc{More training details can be found in the supplementary material.}

\subsubsection{Results}
\mc{
Table~\ref{table:cxr14_2} shows test set performance of ACM compared with other context modules in multiple backbone architectures. ACM achieves the best performance of 85.39 with ResNet-50 and 85.03 with DenseNet121.
We also observe that Non-local (NL) and Cross-Criss Attention (CC) does not perform well in DenseNet architecture, but attains a relatively good performance of 85.08 and 85.11 in ResNet-50. On the other hand, a simpler SE module performs well in DenseNet but does poorly in ResNet50. However, ACM shows consistency across all architectures. One of the possible reasons is that it provides a context based on a contrasting operation, thus unique and helpful across different architectures.
% Given that the CXR14 dataset is a noisy training label dataset~\cite{wang2017chestx} due to its text-mining based labeling method, the general performance improvement across multiple settings is significant.  
}

\begin{table}[t]
\centering
\small
\setlength{\tabcolsep}{3pt}
\caption{\jc{Reported performance of previous works on CXR14 dataset. Each work differs in augmentation schemes and some even in the usage of the dataset. We choose CheXNet~\cite{rajpurkar2017chexnet} as the baseline model for adding context modules.}}
\begin{tabular}{|l|c|c|r|r|}
\hline
Method     & Backbone Arch & Loss Family & Input size   & Reported AUC (\%)   \\ \hline\hline
Wang \etal~\cite{wang2017chestx}        & ResNet-50 & CE &  1,024  & 74.5          \\
Yao \etal~\cite{yao2018weakly}          & ResNet+DenseNet& CE & 512  & 76.1          \\
Wang and Xia~\cite{wang2018chestnet}    &  ResNet-151 & CE & 224 &  78.1          \\
Li \etal~\cite{Li_2018_CVPR}            &  ResNet-v2-50  & BCE  & 299   &    80.6          \\
Guendel \etal~\cite{guendel2018learning} & DenseNet121   & BCE  & 1,024  &   80.7   \\
Guan \etal~\cite{guan2018multi}         & DenseNet121   & BCE  &  224  &   81.6 		 \\
ImageGCN~\cite{mao2019imagegcn}         & Graph Convnet   & CE  & 224   &     82.7 		 \\
CheXNet~\cite{rajpurkar2017chexnet}     & DenseNet121 & BCE & 224 &  84.1  \\       
\hline
\end{tabular}

\label{table:cxr14}
\end{table}

\begin{table}[t]
\centering
\small
\setlength{\tabcolsep}{3pt}
\caption{\mc{Performance in average AUC of various methods on CXR14 dataset. The numbers in the bracket after model names are the input sizes. }}
\begin{tabular}{|l|r|r|}
\hline
Modules        &	DenseNet121(448)  & ResNet-50(448) \\ \hline\hline
None                       & (CheXNet~\cite{rajpurkar2017chexnet}) 84.54  & 84.19 \\
 SE~\cite{hu2018squeeze}    & 84.95  & 84.53\\
 NL~\cite{wang2018non}      & 84.49  & 85.08\\
 CC~\cite{huang2019ccnet}  & 84.43   & 85.11 \\
 ACM                      & \textbf{85.03}  & \textbf{85.39} \\ \hline
\end{tabular}

\label{table:cxr14_2}
\end{table}

\subsection{Detection and Segmentation on COCO}\label{exp:coco}
\subsubsection{Task Overview} In this experiment, we validate the efficacy of ACM in the natural image domain.
Following the previous studies~\cite{wang2018non,huang2019ccnet,cao2019gcnet,woo2018cbam}, we use COCO dataset~\cite{lin2014microsoft} for detection and segmentation tasks. 
Specifically, we use COCO Detection 2017 dataset, which contains more than 200,000 images and 80 object categories with instance-level segmentation masks.
We train both tasks simultaneously using Mask-RCNN architecture~\cite{he2017mask} in Detectron2~\cite{wu2019detectron2}.

\subsubsection{Training Details} 
Basic training details are identical to the default settings in Detectron2~\cite{wu2019detectron2}: learning rate of 0.02 with the batch size of 16. We train for 90,000 iterations, drop the learning rate by 0.1 at iterations 60,000 and 80,000. We use COCO2017-train for training and use COCO2017-val for testing. 

\subsubsection{Results} The experimental results are summarized in Table~\ref{table:coco}. 
Although originally developed for chest X-ray tasks, ACM significantly improves the detection and segmentation performance in the natural image domain as well. In ResNet-50 and ResNeXt-101, ACM outperforms all other modules~\cite{hu2018squeeze,cao2019gcnet,wang2018non}. 
The result implies that the comparing operation is not only crucial for X-ray images but is also generally helpful for parsing scene information in the natural image domain.

\begin{table}[t]
\small
\centering
\setlength{\tabcolsep}{1pt}
%\begin{center}
\caption{Results on COCO dataset. All experiments are based on Mask-RCNN~\cite{he2017mask}.}
\begin{tabular}{|l|ccc|ccc|}
\hline
Method & AP$^\text{bbox}$ & AP$^\text{bbox}_{50}$ & AP$^\text{bbox}_{75}$ & AP$^\text{mask}$ & AP$^\text{mask}_{50}$ & AP$^\text{mask}_{75}$\\
\hline\hline
ResNet-50 & 38.59 & 59.36 & 42.23 & 35.24 & 56.24 & 37.66 \\
ResNet-50+SE~\cite{hu2018squeeze} & 39.10 & 60.32 & 42.59 & 35.72 & 57.16 & 38.20 \\
ResNet-50+NL~\cite{wang2018non} & 39.40 & 60.60 & 43.02 & 35.85 & 57.63 & 38.15 \\
ResNet-50+CC~\cite{huang2019ccnet} & 39.82 & 60.97 & 42.88 & 36.05 & 57.82 & 38.37 \\
ResNet-50+ACM & \textbf{39.94} & \textbf{61.58} & \textbf{43.30} & \textbf{36.40} & \textbf{58.40} & \textbf{38.63} \\
\hline\hline
ResNet-101 & 40.77 & 61.67 & 44.53 & 36.86 & 58.35 & 39.59 \\
ResNet-101+SE~\cite{hu2018squeeze} & 41.30 & 62.36 & 45.26 & 37.38 & 59.34 & 40.00 \\
ResNet-101+NL~\cite{wang2018non} & 41.57 & 62.75 & 45.39 & 37.39 & 59.50 & 40.01 \\
ResNet-101+CC~\cite{huang2019ccnet} & \textbf{42.09} & 63.21 & \textbf{45.79} & \textbf{37.77} & 59.98 & \textbf{40.29} \\
ResNet-101+ACM & 41.76 & \textbf{63.38} & 45.16 & 37.68 & \textbf{60.16} & 40.19 \\
\hline\hline
ResNeXt-101 & 43.23 & 64.42 & 47.47 & 39.02 & 61.10 & 42.11 \\
ResNeXt-101+SE~\cite{hu2018squeeze} & 43.44 & 64.91 & 47.66 & 39.20 & 61.92 & 42.17 \\
ResNeXt-101+NL~\cite{wang2018non} & 43.93 & 65.44 & 48.20 & 39.45 & 61.99 & 42.33 \\
ResNeXt-101+CC~\cite{huang2019ccnet} & 43.86 & 65.28 & 47.74 & 39.26 & 62.06 & 41.97 \\
ResNeXt-101+ACM & \textbf{44.07} & \textbf{65.92} & \textbf{48.33} & \textbf{39.54} & \textbf{62.53} & \textbf{42.44} \\
\hline
\end{tabular}
\label{table:coco}
\end{table}

\subsection{Qualitative Results}
To analyze ACM, we visualize the attention maps for objects of interest and the corresponding context.
The network learns to attend different regions, in such a way to maximize the performance of the given task. 
We visualize the attention maps to see if maximizing performance is aligned with producing attention maps that highlight interpretable locations. Ground-truth annotation contours are also visualized.

We use the Em-Ptx dataset and COCO dataset for the analysis. 
Since there are many attention maps to check, we sort the maps by the amount of overlap between each attention map and the ground truth location of lesions. 
We visualize the attention map with the most overlap. Qualitative results of other tasks are included in the supplementary material due to a space limit.

% ====================================
% qualitative (ptx) figure
\begin{figure}[t]
% \centering
\small
\setlength{\tabcolsep}{1pt}
    \begin{subtable}{0.45\textwidth}
        \begin{tabular}{ccc}
        \includegraphics[height=1.6cm]{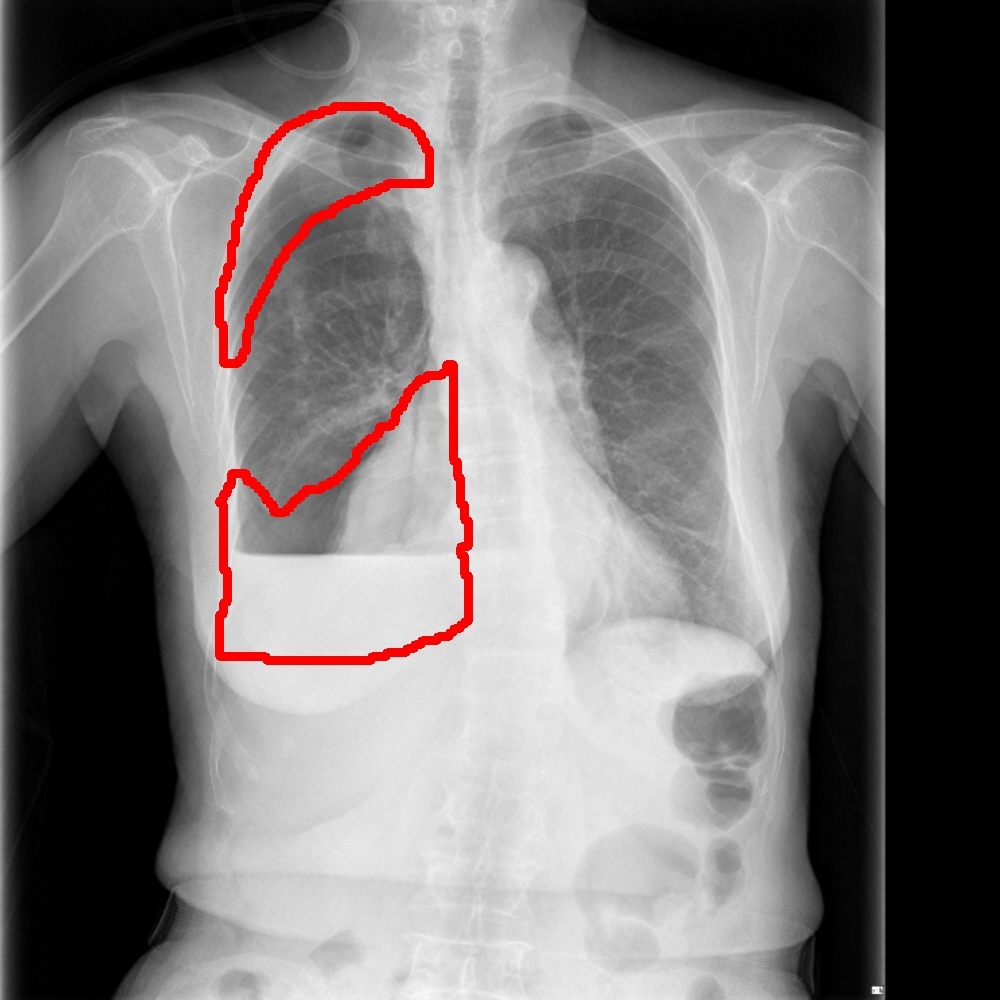}&
        \includegraphics[height=1.6cm]{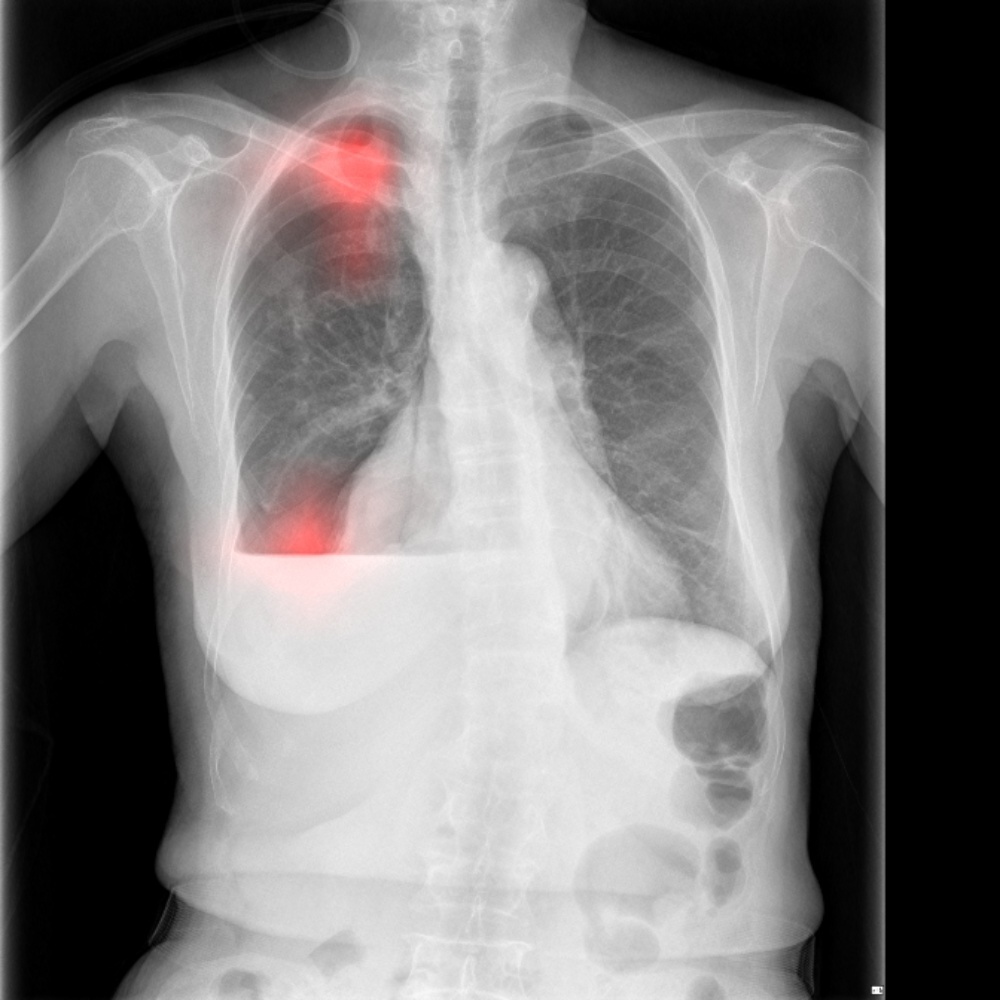}&
        \includegraphics[height=1.6cm]{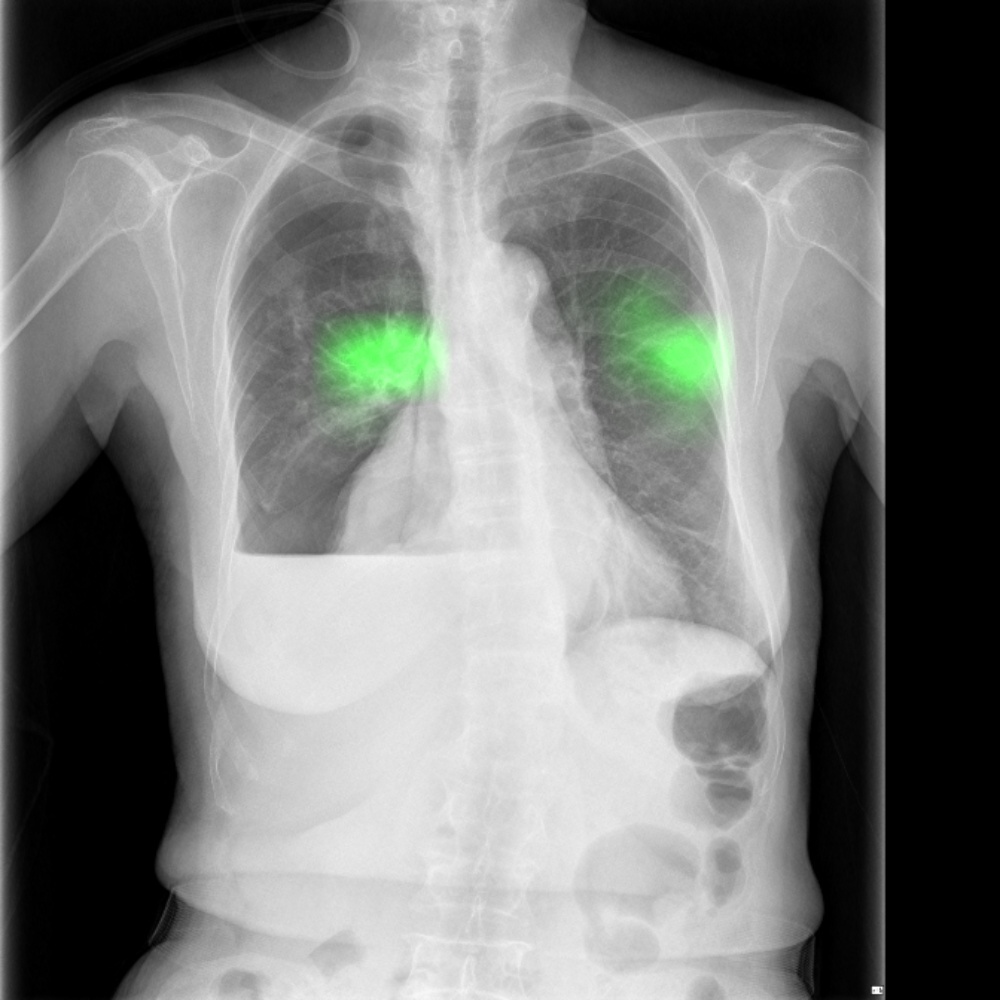}\\
        \includegraphics[height=1.6cm]{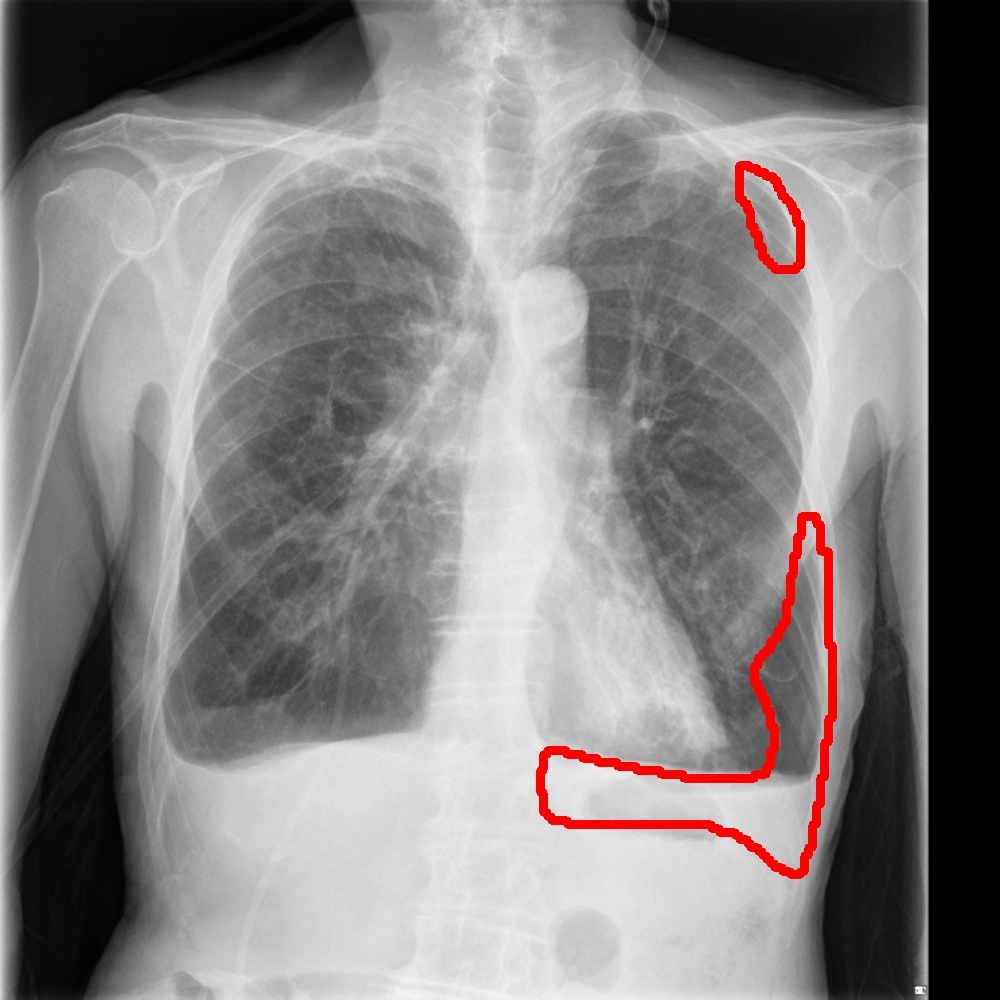}&
        \includegraphics[height=1.6cm]{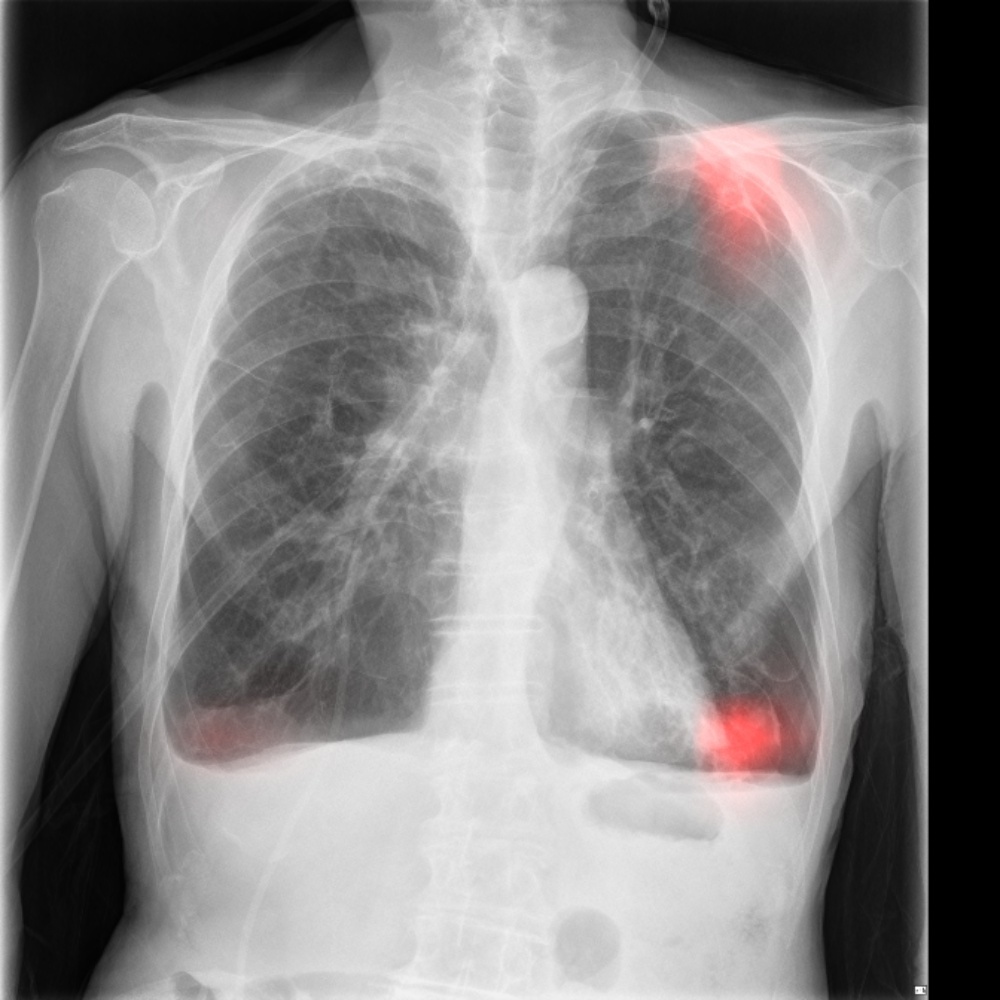}&
        \includegraphics[height=1.6cm]{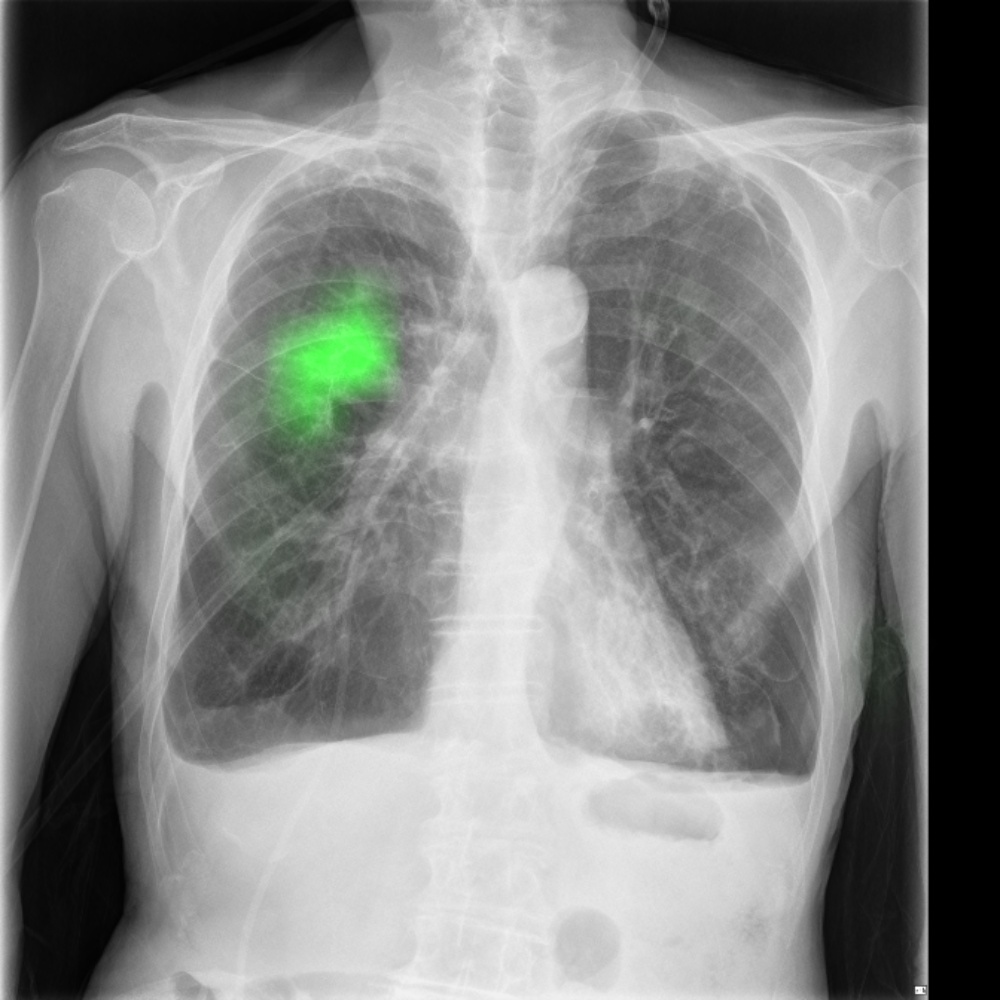}\\
        \includegraphics[height=1.6cm]{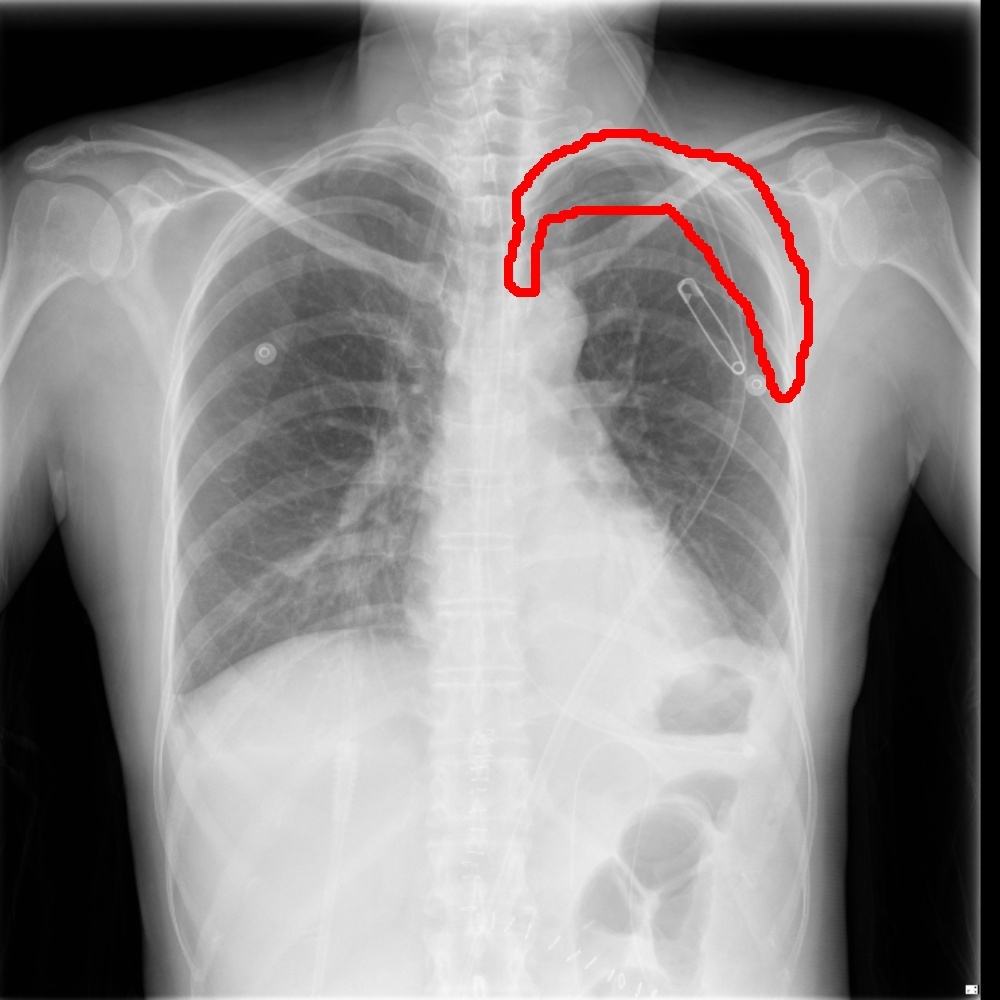}&
        \includegraphics[height=1.6cm]{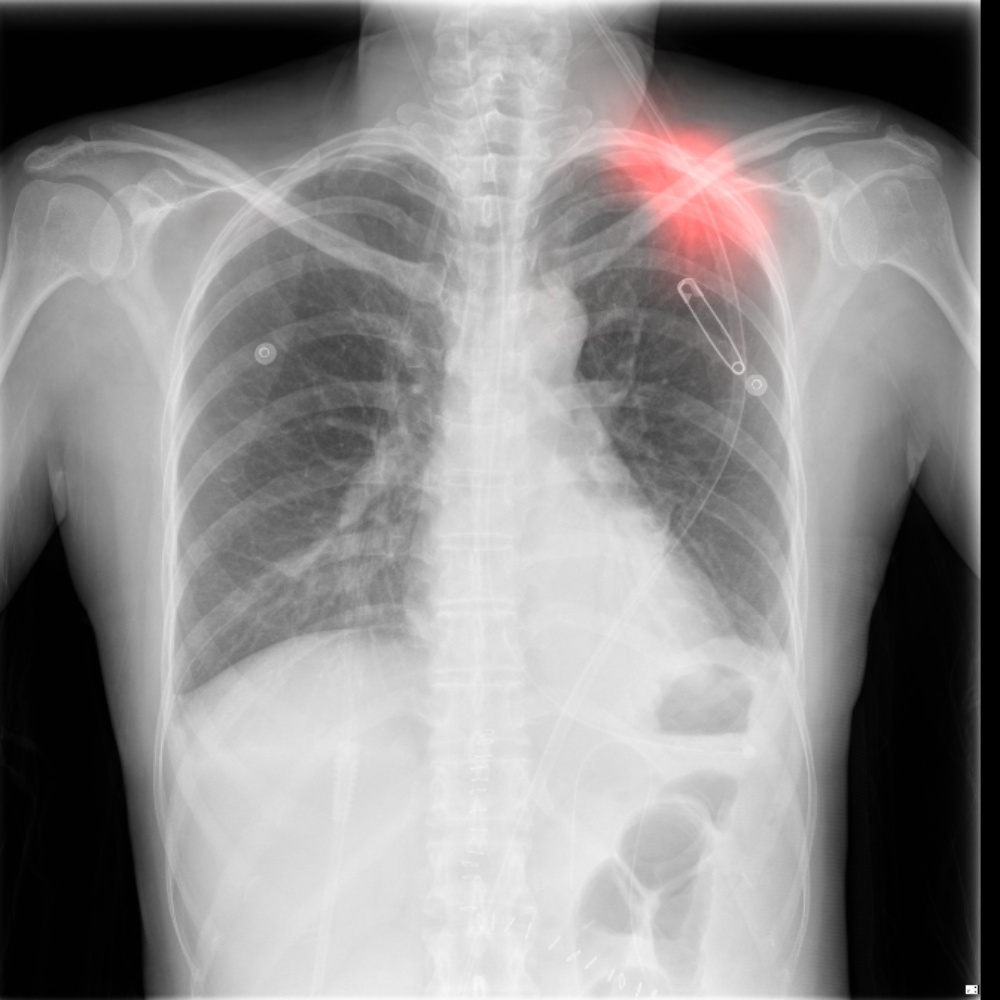}&
        \includegraphics[height=1.6cm]{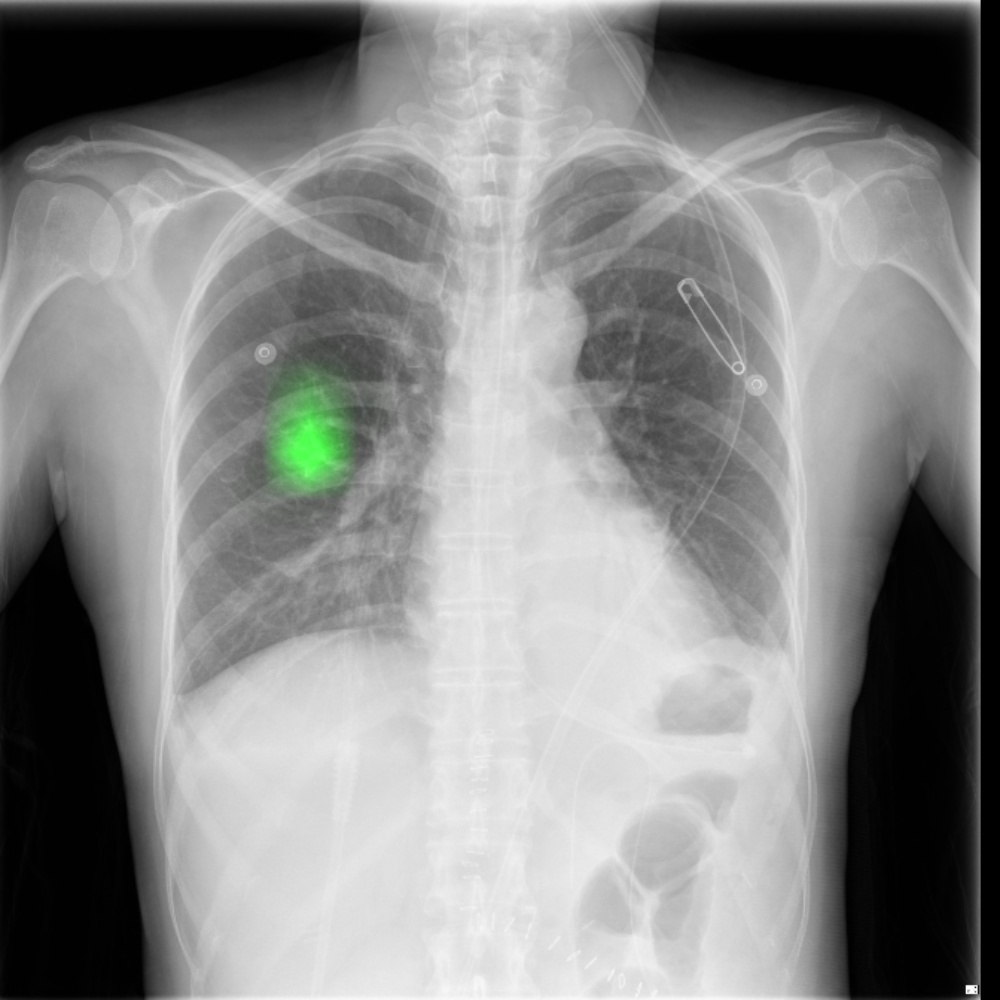}\\
        % Image \& Ptx GT&Att. map for $K$&Att. map for $Q$\\
        GT&$K$ map&$Q$ map\\
        \end{tabular}
    \end{subtable}%
    \begin{subtable}{0.4\textwidth}
        \begin{tabular}{cccc}
        \rotatebox[origin=l]{90}{\;\;Glove}&
        \includegraphics[width=2.0cm]{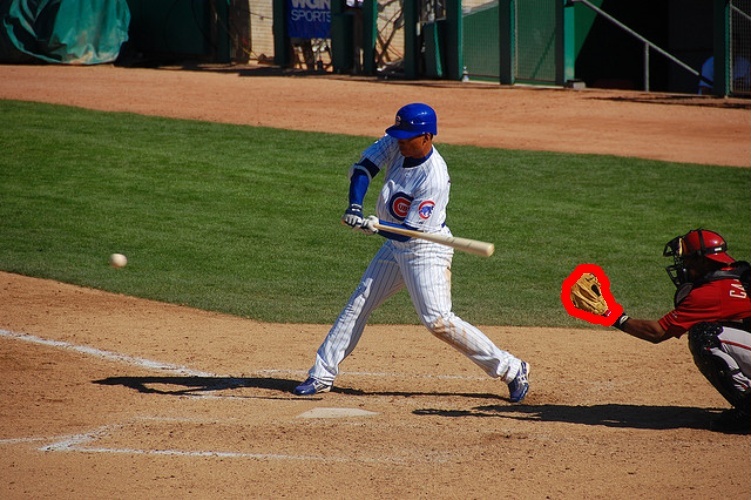}&
        \includegraphics[width=2.0cm]{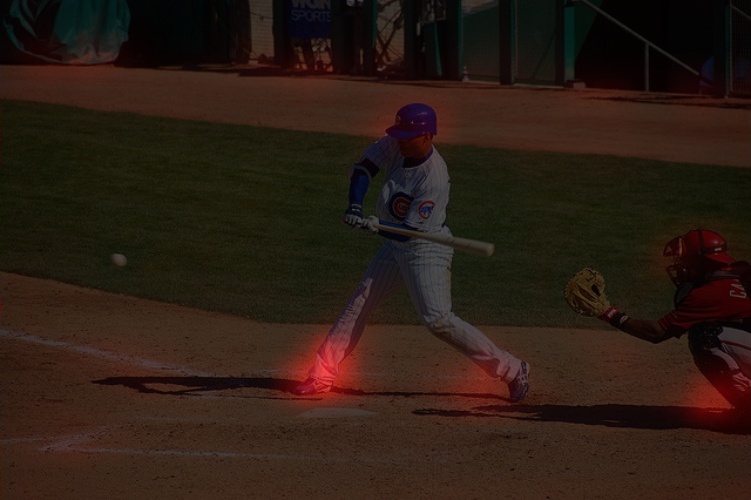}&
        \includegraphics[width=2.0cm]{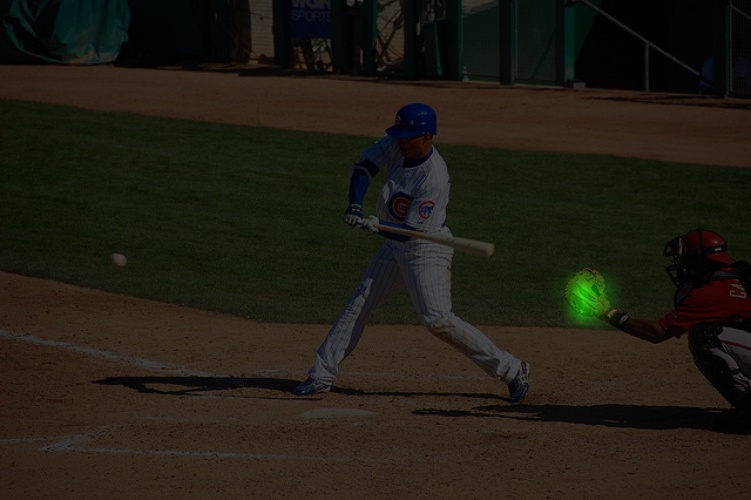}\\
        \rotatebox[origin=l]{90}{\;\;Bicycle}&
        \includegraphics[width=2.0cm]{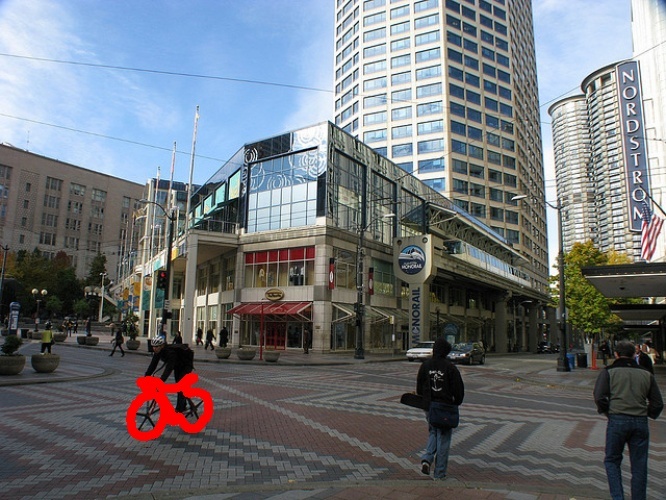}&
        \includegraphics[width=2.0cm]{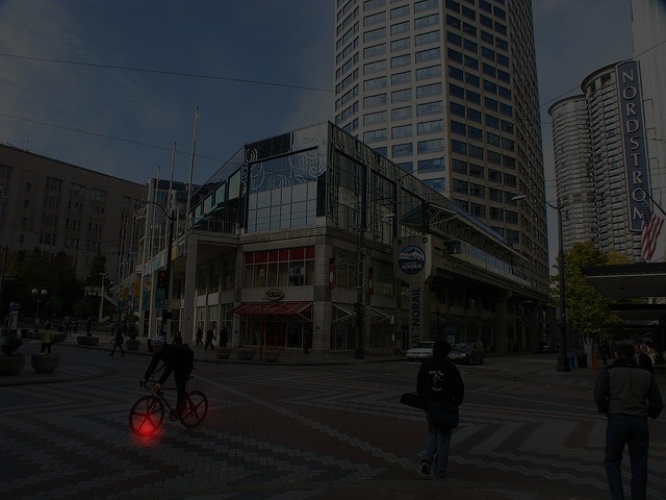}&
        \includegraphics[width=2.0cm]{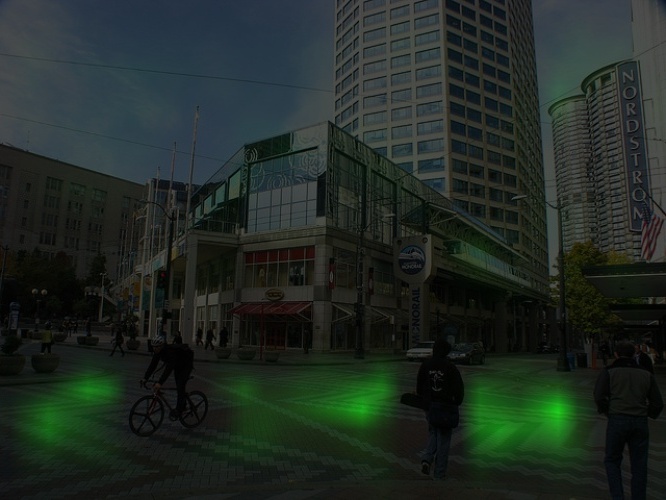}\\
        \rotatebox[origin=l]{90}{\;Frisbee}&
        \includegraphics[width=2.0cm]{figures/qualitative-coco/fris_1_1.jpg}&
        \includegraphics[width=2.0cm]{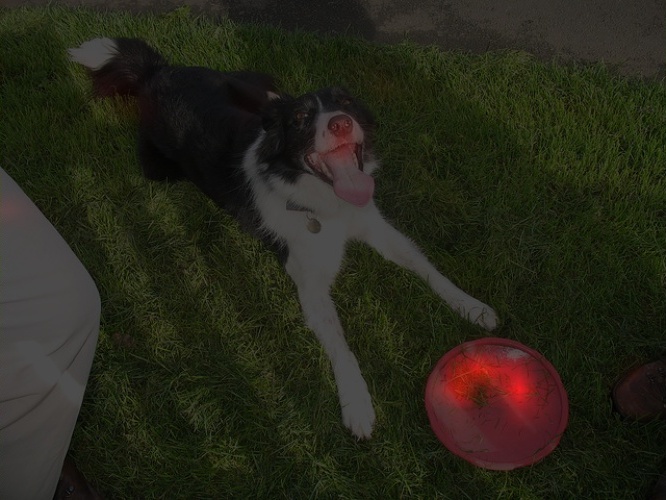}&
        \includegraphics[width=2.0cm]{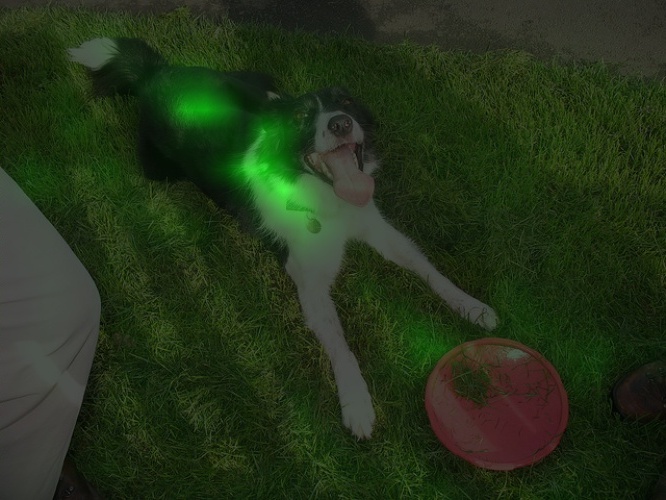}\\
        &GT&$K$ map&$Q$ map\\
        \end{tabular}
    \end{subtable}
\caption{Left: The visualized attention maps for the localization task on Em-Ptx dataset. The 11th group in the 16th module is chosen. Em-Ptx annotations are shown as red contours on the chest X-ray image. 
Right: The visualization on COCO dataset. Ground-truth segmentation annotations for each category are shown as red contours.
}
\label{figure:qualitative_ptx}
\end{figure}

Pneumothorax is a collapsed lung, and on the X-ray image, it is often portrayed as a slightly darker region than the normal side of the lung. 
A simple way to detect pneumothorax is to find a region slightly darker than the normal side.
ACM learns to utilize pneumothorax regions as objects of interest and normal lung regions as the corresponding context. The attention maps are visualized in Fig.~\ref{figure:qualitative_ptx}.
It clearly demonstrates that the module attends to both pneumothorax regions and normal lung regions and compare the two sets of regions. 
The observation is coherent with our intuition that comparing can help recognize, and indicates that ACM automatically learns to do so.

We also visualize the attended regions in the COCO dataset. 
Examples in Fig.~\ref{figure:qualitative_ptx} shows that ACM also learns to utilize the object of interest and the corresponding context in the natural image domain;
for \textit{baseball glove}, ACM combines the corresponding context information from the players' heads and feet; for \textit{bicycle}, ACM combines information from roads; for \textit{frisbee}, ACM combines information from dogs.
We observe that the relationship between the object of interest and the corresponding context is mainly from co-occurring semantics, rather than simply visually similar regions. 
The learned relationship is aligned well with the design principle of ACM; selecting features whose semantics differ, yet whose relationship can serve as meaningful information.

\section{Conclusion}
We have proposed a novel self-contained module, named Attend-and-Compare Module (ACM), whose key idea is to extract an object of interest and a corresponding context and explicitly compare them to make the image representation more distinguishable.
We have empirically validated that ACM indeed improves the performance of visual recognition tasks in chest X-ray and natural image domains. 
Specifically, a simple addition of ACM provides consistent improvements over baselines in COCO as well as Chest X-ray14 public dataset and internally collected Em-Ptx and Ndl dataset. The qualitative analysis shows that ACM automatically learns dynamic relationships. The objects of interest and corresponding contexts are different yet contain useful information for the given task.\jc{The qualitative analysis shows that ACM automatically learns dynamic relationships. The objects of interest and corresponding contexts are different yet contain useful information for the given task.}

\bibliographystyle{splncs04}
\bibliography{egbib}
\end{document}